%% file: main.tex
\documentclass[11pt]{article} 

\input{packages}
\rldmfinaltrue
\input{r}

\title{Moments Matter:\\ Stabilizing Policy Optimization using Return Distributions}

\author{Dennis Jabs\thanks{These authors contributed equally to this work.}\,\\
Leibniz University Hannover\\
\texttt{dennis-jabs@outlook.de} \\
\And 
Aditya Mohan \footnotemark[1]\,\\
Intitute of Artificial Intelligence \\
Leibniz University Hannover \\
\texttt{a.mohan@ai.uni-hannover.de} \\
\And 
Marius Lindauer \\
Intitute of Artificial Intelligence\\ 
L3S Research Center \\
Leibniz University Hannover\\
}

\begin{document}

\maketitle

\input{Contents/Abstract}
\input{Contents/Introduction}
\input{Contents/Preliminaries}
\input{Contents/Alignment}
\input{Contents/Distirbutions}
\input{Contents/Experiments}

\section*{Acknowledgements}
The authors acknowledge funding by the German Research Foundation (DFG): Aditya Mohan under LI 2801/7-1.

\bibliography{bib/shortstrings, bib/lib,bib/local,bib/shortproc}
\end{document}

%% file: packages.tex
\usepackage{rldmsubmit,palatino}
\usepackage{graphicx}

\usepackage{algorithm}
\usepackage{algorithmic}

\usepackage{nicefrac}
\usepackage[utf8]{inputenc}
\usepackage{booktabs}
\usepackage{amsfonts}

\usepackage{amsmath}
\usepackage{amssymb}
\usepackage{mathtools}
\usepackage{amsthm}
\newtheorem{theorem}{Hypothesis}
\usepackage{siunitx}
\usepackage{fnpct}

\usepackage{microtype}
\usepackage{graphicx}
\usepackage{booktabs} 

\usepackage{hyperref}
\hypersetup{
    colorlinks,
    linkcolor={red},
    citecolor={blue},
    urlcolor={blue}
}
\usepackage{url}
\usepackage[capitalize,noabbrev]{cleveref}

\usepackage{xspace}
\usepackage{placeins}
\usepackage{tabularx}
\usepackage{pifont}
\usepackage[nolist]{acronym}
\usepackage{subcaption}

\usepackage{tikz}
\usepackage{caption}
\usepackage{subcaption}
\usetikzlibrary{positioning, arrows, shapes,snakes, calc}
\usetikzlibrary{shadows,shadings,shapes.symbols}

\usepackage{wrapfig}
\usepackage{placeins}

\usepackage{paralist}
\usepackage{bm}

\usepackage{cancel}

\usepackage{natbib}
\usepackage{makecell}

%% file: Contents/Abstract.tex
\begin{abstract}
Deep Reinforcement Learning (RL) agents often learn policies that achieve the same episodic return yet behave very differently, due to a combination of environmental (random transitions, initial conditions, reward noise) and algorithmic (minibatch selection, exploration noise) factors. 
In continuous control tasks, even small parameter shifts can produce unstable gaits, complicating both algorithm comparison and real-world transfer. 
Previous work has shown that such instability arises when policy updates traverse noisy neighborhoods and that the spread of \emph{post-update return distribution} $\mathcal{R}(\theta)$ -- obtained by repeatedly sampling minibatches, updating $\theta$, and measuring final returns -- is a useful indicator of this noise. 
Although explicitly constraining the policy to maintain a narrow $\mathcal{R}(\theta)$ can improve stability, directly estimating $\mathcal{R}(\theta)$ is computationally expensive in high-dimensional settings.
We propose an alternative that takes advantage of \emph{environmental stochasticity} to mitigate update-induced variability. Specifically, we model \emph{state-action return distribution} through a distributional critic and then bias the advantage function of PPO using higher-order moments (skewness and kurtosis) of this distribution. 
By penalizing extreme tail behaviors, our method discourages policies from entering parameter regimes prone to instability. 
We hypothesize that in environments where post-update critic values align poorly with post-update returns, standard PPO struggles to produce a narrow $\mathcal{R}(\theta)$. 
In such cases, our moment-based correction narrows $\mathcal{R}(\theta)$, improving stability by up to 75\% in Walker2D, while preserving comparable evaluation returns.
\end{abstract}

%% file: Contents/Introduction.tex
\vspace{-4pt}
\section{Introduction}
Agents trained with Deep Reinforcement Learning (RL) often exhibit significant variability in performance, such as sudden and unpredictable drops in returns, even under seemingly identical conditions. 
This complicates reliable progress measurement and algorithm comparison. 
In continuous control settings, policies can exhibit divergent behaviors even if they get the same evaluation returns, making deployment to the real world challenging.
This variability arises not only from environmental stochasticity but also from sensitivity to small policy perturbations, where slight changes in policy parameters can lead to disproportionately large changes in behavior.
Consequently, maintaining stable performance across slight policy perturbations remains a core challenge in RL.

\emph{Environmental Stochasticity}, caused by random state transitions, initial conditions, and reward noise in sampled trajectories, affects the returns collected during learning. 
\emph{Algorithmic Stochasticity}, arising from minibatch sampling and random exploration during optimization, influences how policy parameters change over iterations.
Together, they form a closed-loop feedback system: environment-induced variability influences the mini-batches used for updates, while algorithmic stochasticity alters the distribution of visited states in potentially unpredictable ways.

A direct way to measure such instabilities is by constructing the \emph{post-update return distribution} $\mathcal{R}(\theta)$,
the distribution of returns obtained after applying updates to policy parameters.
The tail of $\mathcal{R}(\theta)$ indicates the frequency of policy updates that lead to low returns, and explicitly rejecting updates that produce larger tails reduces the frequency of catastrophic failures, offering a clearer path to stable learning \citep{rahn-neurips23a}.
However, estimating $\mathcal{R}(\theta)$ requires repeated Monte Carlo evaluations of multiple policy updates, which quickly becomes intractable in high-dimensional or long-horizon tasks.

We propose an alternative that takes advantage of environmental stochasticity. 
First, we hypothesize that a key factor driving stochasticity in policy updates is the misalignment between the predicted values of the critic and the accumulated returns (\Cref{sec:alignment}). 
We then propose a correction mechanism for such environments by \emph{replacing the single-value critic} with \emph{distributional critic} \citep{bellemare-icml17a,schneider-icra24a} and using its higher order properties (skewness, kurtosis) as regularization terms in advantage estimation (\Cref{sec:distributions}). 
By incorporating these statistics into the update rule, agents can proactively steer policy parameters toward smoother, more reliable performance. 
Empirically, we demonstrate that our framework improves stability in contrinous control tasks -- measured by the narrowness of $\mathcal{R}(\theta)$ -- by up to $75\%$ in Walker2D, while preserving the evaluation returns (\Cref{sec:experiments}).

%% file: Contents/Preliminaries.tex
\section{Preliminaries}

We consider a discounted continuous Markov Decision Process (MDP) $\mathcal{M} = (\mathcal{S}, \mathcal{A}, p, r, \gamma, \rho_0)$.
At each time step $t$, the agent observes a state $s_t \in \mathcal{S}$, selects an action $a_t \in \mathcal{A}$ according to a stochastic policy $\pi_\theta(a \mid s)$ parameterized by $\theta$, transitions to a new state $s_{t+1} \sim p(\cdot \mid s_t, a_t)$, and receives a reward $r(s_t, a_t)$. The objective is to maximize the expected return, defined in the episodic setting as: $R_t = \sum_{k=0}^{T} \gamma^k r(s_{t+k}, a_{t+k}),$ where $T$ is the (finite) episode length, and the expectation is taken over the stochastic dynamics of the environment and the policy.

\textbf{State-Action Return Distribution.}  
Even for a fixed policy, randomness in state transitions and rewards, due to the stochastic nature of $p(s' \mid s, a)$ and $r(s, a)$, results in a distribution of returns from any given state-action pair. 
This distribution is referred to as the \emph{state-action return distribution} \( Z^\pi(s, a) \), which captures the variability in the outcomes caused by environmental stochasticity. 
A distributional critic \( Z_\phi(s, a) \) parameterized by $\phi$ can be trained to approximate this distribution~\citep{bellemare-icml17a}. 

\textbf{Post-Update Return Distribution.}  
In contrast to \( Z^\pi(s, a) \), \emph{post-update return distribution} $\mathcal{R}(\theta)$ captures the variability in returns caused by the learning process itself. 
This arises from two primary sources of randomness during policy optimization.
\begin{inparaenum}[(i)]
    \item \textit{Minibatch sampling} during gradient-based updates; and
    \item \textit{Exploration noise} added to actions to encourage exploration during training.
\end{inparaenum}
These factors lead to variability in the sequence of policy parameters \( \{\theta_t\} \) over time and, consequently, in the returns generated by the updated policies.
We focus on minibatches as the sources of stochasticity by setting the exploration hyperparameter off.
Formally, Given $N$ minibatches $\{X_i\}_{i=1}^{N}$, we generate a set of updated policy parameters 
$\{\theta_i' = U(\theta, X_i)\}_{i=1}^{N},$  where $U(\theta, X_i)$ denotes an update rule applied to the current policy parameters $\theta$ using data from minibatch $X_i$. 
For each updated policy $\theta_i'$, we evaluate its performance in the environment to obtain a return $R_i$ and thus create the \emph{post-update return distribution} $\mathcal{R}(\theta_i')$.

%% file: Contents/Alignment.tex
\section{Alignment Between Critic and Return}
\label{sec:alignment}
 

In policy optimization algorithms such as Proximal Policy Optimization (PPO; \citealp{schulman-arxiv17a}), the critic baseline guides policy updates by estimating expected returns. 
However, each update can move the policy in ways that introduce significant variability in the actual returns. 
Complex or highly stochastic environment dynamics can further exacerbate this effect, making it difficult for the critic to accurately capture the resulting range of outcomes. 
To detect when the critic’s estimates deviate from the observed returns, we propose monitoring both the post-update return distribution and the predicted value distribution of the critic, then comparing their variability to assess the critic's fidelity to tracking the true returns.

Given $N$ minibatches $\{X_i\}_{i=1}^{N}$, in addition to recording the returns $R_i'$ for each parameter $\theta_i'$ generated after applying the update $\{\theta_i' = U(\theta, X_i)\}_{i=1}^{N},$ we additionally record the baseline estimates of the critic $V_{\theta_i'}(s)$ for the same minibatches to form the \emph{Post-update Value Distribution}:
$
   \mathcal{V}(\theta) 
   \;=\; 
   \{
       V_{\theta_i'}(s)
       \; \mid \;
       \theta_i' = U(\theta, X_i),\; s \in S
   \}_{i=1}^N.
$

\paragraph{Tracking Variability for Stability.}
High variance in $\mathcal{R}(\theta)$ indicates that policy updates produce large performance swings, potentially affecting the reliability of advantage estimates. 
If the critic \emph{also} exhibits a high variance $\mathcal{V}(\theta)$, then it is likely to track this observed volatility, allowing the policy to adapt more cautiously (e.g., through tighter clipping or lower learning rates). 
In contrast, if the critic's variance remains low while the actual returns fluctuate widely, the critic’s estimates may be misleading, leading to erratic updates. Monitoring the correlation between these variances over multiple post-update distributions (rather than relying on a single snapshot) helps reveal longer-term trends and can prompt adjustments that stabilize PPO.

\begin{theorem}[Weak Alignment Causes Instability]
\label{thm:weak-alignment}
In environments where the post-update value distribution $\mathcal{V}(\theta)$ aligns weakly with the post-update return distribution $\mathcal{R}(\theta)$, the PPO algorithm is prone to erratic performance and unstable policy updates due to unreliable advantage estimates.
\end{theorem}

Mathematically, let $\sigma^2(\mathcal{R}(\theta))$ denote the variance of the post-update return distribution and $\sigma^2(\mathcal{V}(\theta))$ the variance of the post-update value distribution. When the critic accurately tracks the variability in returns, we expect these two variances to be strongly correlated:
$
   r 
   \;=\; 
   \text{Corr}\bigl(\sigma^2(\mathcal{R}(\theta)),\;\sigma^2(\mathcal{V}(\theta))\bigr)
   \;\approx\;
   1.
$
A low correlation suggests that the critic’s variance estimate is not representative of actual returns, leading to misleading advantage estimates and potentially erratic policy updates. Ensuring a high degree of alignment between these two distributions helps to maintain more stable learning dynamics in PPO.

%% file: Contents/Distirbutions.tex
\section{Stabilizing Policy Updates using a Distributional Critic}
\label{sec:distributions}

\begin{figure}[!htb]
    \centering
    \begin{minipage}[t]{0.45\textwidth}
        \centering
        \includegraphics[width=0.9\textwidth]{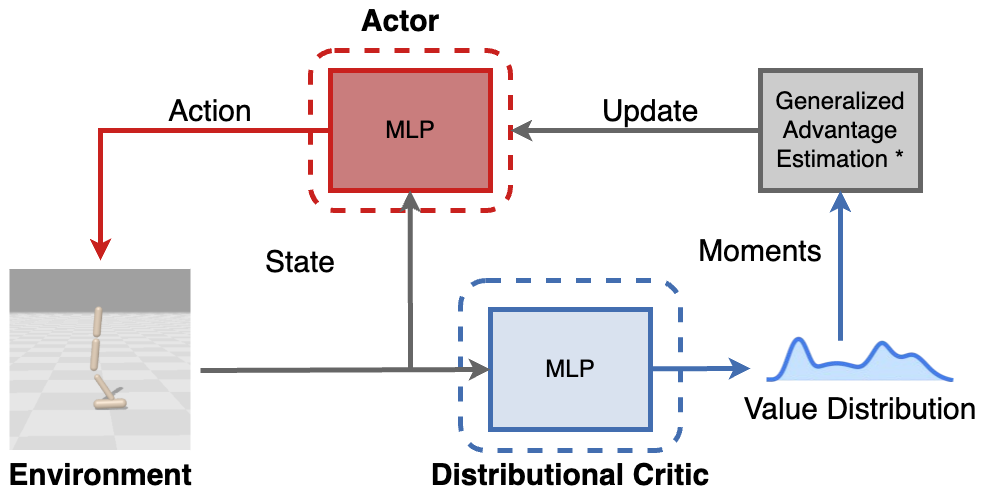}
        \caption{Overview of the architecture of PPO. We modify the critic to a distributional critic that uses a $51$-quantile support distribution as outlined by \cite{schneider-icra24a} to approximate the return distribution. We then utilize the higher-order moments of this distribution to regularize the advantage estimation.}
        \label{fig:setup}
    \end{minipage}
    \hfill
    \begin{minipage}[t]{0.45\textwidth}
        \centering
        \vspace{-3cm} 
        \begin{tabular}{lcc}
            \toprule
            Environment & Pearson corr. coef. $r$ & p-value $\rho$ \\
            \midrule
            Ant         & 0.49 & 0.0 \\
            HalfCheetah & \textbf{0.74} & 0.0 \\
            Hopper      & 0.50 & 0.0 \\
            Walker2D    & 0.20 & 0.0 \\
            \bottomrule
            \vspace{1mm}
        \end{tabular}
        \captionof{table}{Pearson correlation coefficients $r$ and their p-values $\rho$ computed under the null hypothesis $H_0$ of no linear relationship between $\mathcal{V}(\theta)$ and $\mathcal{R}(\theta)$. A higher $r$ indicates stronger alignment between the critic and the return, while $\rho = 0.0$ indicate that all observed correlations are statistically significant.}
        \label{tab:pearson}
    \end{minipage}
\end{figure}

In environments with high stochasticity or complex dynamics, a single-point estimate of the expected return can fail to capture the overall risk, producing policy updates that are overly aggressive in high-variance regions or overly conservative in regions with skewed or heavy-tailed distributions. 
A distributional critic, which models the full return distribution, offers a more complete representation of future outcomes. 
Higher-order properties of such distributions, such as skewness and kurtosis, can guide policy updates to more stable regions, thereby improving learning stability and performance.

\vspace{-1pt}
\begin{theorem}[Higher-Order Moments Stabilize Misaligned Updates]
In environments with high variability, penalizing excess skewness and kurtosis biases policy updates away from regions with extreme asymmetry or heavy-tailed distributions, reducing the likelihood of erratic returns and enhancing stability.
\end{theorem}

\textbf{Distributional Critic Setup.}
Figure~\ref{fig:setup} shows our distribution set-up using DPPO~\citep{schneider-icra24a}:
We replace the critic in PPO with a distributional critic $Z_\phi(s,a)$ that uses a quantile-based approximation to model the full return distribution under policy $\pi_\theta$. 
We then capture higher-order momenets of $Z_\phi(s,a)$, specifically:
\begin{inparaenum}[(i)]
    \item \textbf{Skewness}: Represents the asymmetry of the distribution, $\text{Skew}(Z_{\phi}) \;=\; \frac{\mathbb{E}\bigl[(Z_{\phi} - \mathbb{E}[Z_{\phi}])^3\bigr]}{\sigma^3(Z_{\phi})} $. 
    A high absolute skewness indicates a strong imbalance, where one side of the distribution contains more extreme outcomes, even if the expected value remains unchanged.
    \item \textbf{Kurtosis}: Quantifies the tailedness of the distribution $\text{Kurt}(Z_{\phi}) \;=\; \frac{\mathbb{E}\bigl[(Z_{\phi} - \mathbb{E}[Z_{\phi}])^4\bigr]}{\sigma^4(Z_{\phi})}.$. High kurtosis signals the presence of heavy tails, implying more frequent extreme outcomes, which can disrupt stable learning.
    Negative tails are generally undesirable, but even extremely good outliers are still hard to learn and attribute.
\end{inparaenum}

\textbf{Regularizing for Skewness and Kurtosis.}
We bias the advantage estimation mechanism $\hat{A}_t$, to reduce the occurrence of asymmetric or heavy-tailed value distributions. 
Excess skewness is penalized to mitigate asymmetry, while excess kurtosis is penalized to suppress extreme outliers. 
The penalization terms are shown in Eq.~\ref{eq:regularization}, where $w_{\text{skew}}, w_{\text{kurt}} \geq 0$ are hyperparameters that control the penalization strengths.
\begin{equation}
\label{eq:regularization}
\hat{A}_t \leftarrow \hat{A}_t - w_{\text{skew}} \text{Skew}(Z_\phi(s_t)),
\qquad \qquad
\hat{A}_t \leftarrow \hat{A}_t - w_{\text{kurt}} \text{Kurt}(Z_\phi(s_t)),
\end{equation}

%% file: Contents/Experiments.tex
\section{Empirical Validation of Higher-Order Moment Penalization}
\label{sec:experiments}

\textbf{Setup.}
We conducted experiments on four continuous control tasks from Brax: \emph{Ant}, \emph{HalfCheetah}, \emph{Hopper}, and \emph{Walker2D}. Each agent was trained for $6 \times 10^7$ time steps using 10 random seeds. We implemented \emph{PPO} \citep{schulman-arxiv17a}, post-update CVaR rejection sampling (\emph{CRS}) \citep{rahn-neurips23a}, \emph{DPPO} \citep{schneider-icra24a}, and our regularized variants \emph{DPPO-Kurt} and \emph{DPPO-Skew} (penalizing kurtosis and skewness, respectively) using PureJaxRL~\citep{lu-neurips22a}.

\begin{figure}[htbp]
    \centering
    \begin{minipage}{0.7\textwidth}
        \includegraphics[width=\textwidth]{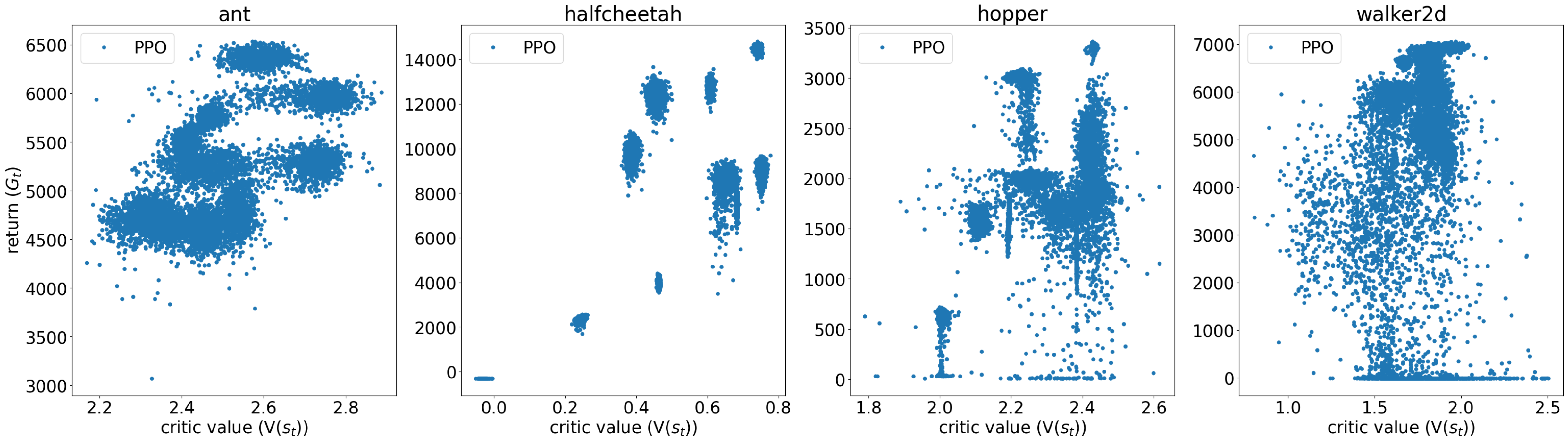}
    \end{minipage}%
    \hspace{1em} 
    \begin{minipage}{0.25\textwidth}
        \caption{Relationship between returns and critic values sampled from $\mathcal{R}(\theta)$ and $\mathcal{V}(\theta)$ of PPO using the final checkpoint across Brax environments.}
        \label{fig:alignment}
    \end{minipage}
\end{figure}

\textbf{Measuring Alignment.} 
To measure the alignment between $\mathcal{R}(\theta)$ and $\mathcal{V}(\theta)$, we sampled $10{,}000$ minibatches for each environment, generated updated policies and recorded the returns and the corresponding critic values.
Figure~\ref{fig:alignment} plots these relationships, and Table~\ref{tab:pearson} summarizes their Pearson correlation. 
Among the four tasks, only \textit{HalfCheetah} shows a relatively strong relationship between $\mathcal{R}(\theta)$ and $\mathcal{V}(\theta)$, while the other environments show significant scatter, indicating poor critic-return alignment.

\begin{figure}[htb]
    \centering
    \includegraphics[width=0.9\textwidth,trim={0 1.5cm 0 0},clip]{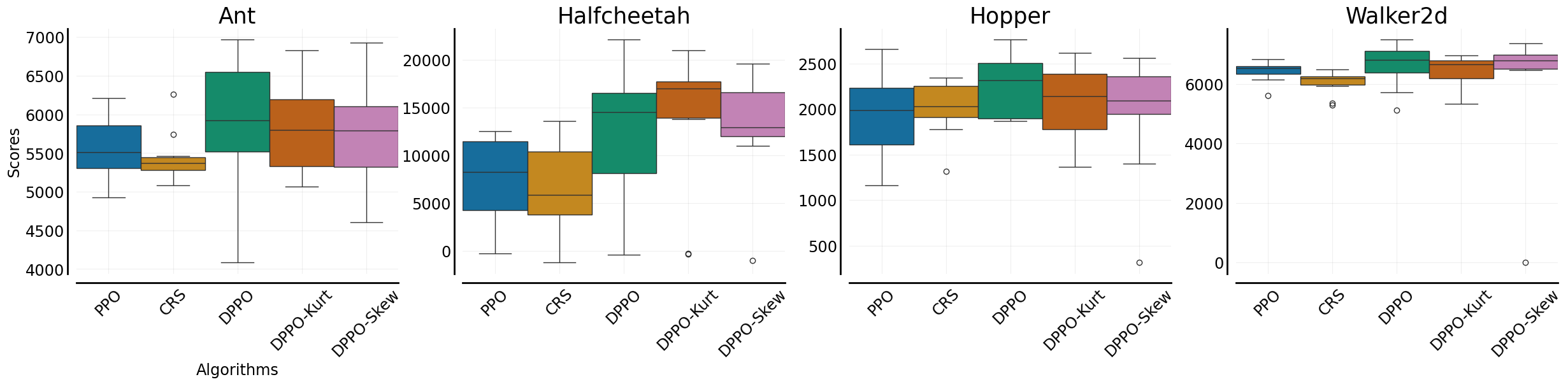}
    \caption{Final evaluation returns across $10$ seeds for Brax environments.}
    \label{fig:boxplots}
\end{figure}

\textbf{Performance and Stability Evaluation.}
We next evaluated each algorithm’s training performance (Figure~\ref{fig:boxplots}) and measured stability by computing the standard deviation $\sigma$ of $\mathcal{R}(\theta)$ (Table~\ref{tab:std-dev-results}).
In tasks with weak alignment (e.g., \emph{Ant}, \emph{Hopper}, \emph{Walker2D}), our distributionally regularized variants (\emph{DPPO-Kurt}, \emph{DPPO-Skew}) produce narrower $\mathcal{R}(\theta)$ at similar return levels. 
Meanwhile, in \textit{HalfCheetah} -- the only task that shows strong critic-return alignment -- the standard PPO already yields the narrowest $\mathcal{R}(\theta)$. These findings directly support \emph{Hypothesis 1} and underscore that vanilla PPO performs well when critic and return are strongly correlated.

\textbf{Impact of Higher-Order Moment Penalization.}
Comparing \emph{DPPO-Kurt} and \emph{DPPO-Skew} to \emph{DPPO} further clarifies that simply modeling the distribution of returns is insufficient; \emph{explicitly} regularizing higher order moments is the key to narrowing $\mathcal{R}(\theta)$. 
Moreover, the fact that these regularized variants produce a tighter $\mathcal{R}(\theta)$ than \emph{CRS} highlights the advantages of our approach over the explicit MC-based estimation of $\mathcal{R}(\theta)$. 
Finally, \emph{DPPO-Kurt} often produces a narrower distribution than \emph{DPPO-Skew}, suggesting that kurtosis regularization may be particularly beneficial for stability.

\begin{figure}[h!]
    \centering
    \begin{minipage}[t]{0.5\textwidth}
        \centering
        \begin{tabular}{lccccc}
            \toprule
            \textbf{Environment} & \textbf{PPO} & \textbf{CRS} & \textbf{DPPO} & \textbf{DPPO-Kurt} & \textbf{DPPO-Skew} \\
            \midrule
            Ant           & 105.51            & 89.06               & 94.03            & \textbf{77.07}    & 98.23            \\
            HalfCheetah   & \textbf{264.56}   & 369.67              & 312.81           & 413.20            & 287.62           \\
            Hopper        & 234.31            & 207.13              & 233.87           & \textbf{57.75}    & 142.34           \\
            Walker2D      & 799.44            & 643.28              & 1246.18          & \textbf{196.06}   & 226.82           \\
            \bottomrule
        \end{tabular}
    \end{minipage}
    \hfill
    \begin{minipage}[t]{0.3\textwidth}
        \vspace{-1.2cm}
        \captionof{table}{Standard deviation $\sigma$ of $\mathcal{R}(\theta)$, averaged over $10$ seeds, of policies produced by different algorithms across continuous control tasks. The lowest $\sigma$ in each environment is highlighted.}
        \label{tab:std-dev-results}
    \end{minipage}
\end{figure}

\vspace{-5pt}

\textbf{Conclusion.}
Our experiments show that when the critic aligns closely with the post-update returns (as in \emph{HalfCheetah}), the vanilla PPO already maintains a narrow $\mathcal{R}(\theta)$. However, in tasks that exhibit poor alignment, our distributional moment regularization narrows $\mathcal{R}(\theta)$ by up to 75\% in \emph{Walker2D} with \emph{DPPO-Kurt}, substantially improving stability while retaining competitive performance. 
Future work includes: 
\begin{inparaenum}[(i)]
    \item developing a formal theoretical framework for our two hypotheses, 
    \item exploring the connectivity of policy optimization paths, and 
    \item extending these methods to broader domains.
\end{inparaenum}